\documentclass[runningheads]{llncs}
\usepackage{graphicx}
\usepackage{algorithmic}
\usepackage{multirow}
\usepackage{graphicx}
\usepackage{newtxtext}
\usepackage[misc]{ifsym}
\usepackage{amsmath}
\usepackage{cite}
\usepackage{textcomp}
\usepackage[ruled,vlined]{algorithm2e}
\usepackage{xcolor}
\usepackage{amsmath,amssymb,amsfonts}

\usepackage{url}

\begin{document}
\title{Frustratingly Easy Feature Reconstruction for Out-of-Distribution Detection}

\author{
  Yingsheng Wang\inst{1,2} \and
  Shuo Lu\inst{1,2} \and
  Jian Liang\inst{2}$^{(\textrm{\Letter})}$ \and
  Aihua Zheng\inst{3} \and
  Ran He\inst{4}}
\authorrunning{Y. Wang et al.}
\institute{
  MAIS \& NLPR, Institute of Automation, Chinese Academy of Sciences, Beijing, China \and
  School of Artificial Intelligence, University of Chinese Academy of Sciences, Beijing, China \and
  Anhui University, Hefei, China \\
  \email{wa23301185@stu.ahu.edu.cn}, \email{\{shuolucs, liangjian92\}@gmail.com},  \email{ahzheng214@foxmail.com}, \email{rhe@nlpr.ia.ac.cn}
}

\maketitle              
\begin{abstract}
Out-of-distribution (OOD) detection helps models identify data outside the training categories, crucial for security applications. While feature-based post-hoc methods address this by evaluating data differences in the feature space without changing network parameters, they often require access to training data, which may not be suitable for some data privacy scenarios. This may not be suitable in scenarios where data privacy protection is a concern. In this paper, we propose a simple yet effective post-hoc method, termed Classifier-based Feature Reconstruction (ClaFR), from the perspective of subspace projection. It first performs an orthogonal decomposition of the classifier's weights to extract the class-known subspace, then maps the original data features into this subspace to obtain new data representations. Subsequently, the OOD score is determined by calculating the feature reconstruction error of the data within the subspace. Compared to existing OOD detection algorithms, our method does not require access to training data while achieving leading performance on multiple OOD benchmarks. 
Our code is released at \url{https://github.com/Aie0923/ClaFR}.

\end{abstract}
\section{Introduction}
\label{sec:intro}
When deploying deep neural networks in the open world, discrepancies between the training and test data distributions often lead to the model overconfidence, which can result in significant flaws during deployment~\cite{hein2019relu, dosovitskiy2020image, liang2023realistic, he2010two}.
This issue is particularly critical in safety-critical domains such as autonomous driving~\cite{chen2017no, liu2021visual} and medical image analysis~\cite{liu2022source, Daniel2020Generalized}.
Current research in OOD detection focuses on enabling models to recognize their limitations and reject predictions with low confidence, thus addressing the problem of model overconfidence on unfamiliar samples~\cite{yang2022openood, pu2021anomaly, chan2021entropy, yang2023auto, yang2024generalized}.
The OOD detection task primarily involves creating a scoring function that assigns a confidence score to each sample, indicating its likelihood of being ID sample, and using a threshold to help the model differentiate between data types.

Many existing Out-of-Distribution (OOD) detection methods focus on analyzing feature variations within neural networks, as evidenced by works from ~\cite{sun2022dice, ming2022exploit, yang2023auto, lu2024learning}. Among these, a significant category comprises training-agnostic feature-based methods~\cite{huang2021importance, morteza2022provable, park2023nearest, zhao2024towards, peng2024conjnorm}. These approaches specifically emphasize feature shaping and analysis, typically by processing features extracted from the neural network to amplify the inherent differences between ID and OOD data features, all without requiring further model retraining.
ViM~\cite{wang2022vim} leverages subspace projection to obtain residuals for calculating virtual logits, effectively combining information from both the feature space and logits to aid distinction.
~\cite{guan2023revisit} involves using subspace to compute the reconstruction error of features.
These methods operate under the fundamental assumption that ID sample features exhibit a compact distribution aligned with the principal components within the feature space. Conversely, OOD data, due to the model's lack of prior exposure and fitting to their characteristics, cannot guarantee such a clustered or aligned distribution, making them distinguishable.

\begin{figure}[h!]
  \centering
  \includegraphics[width=\linewidth]{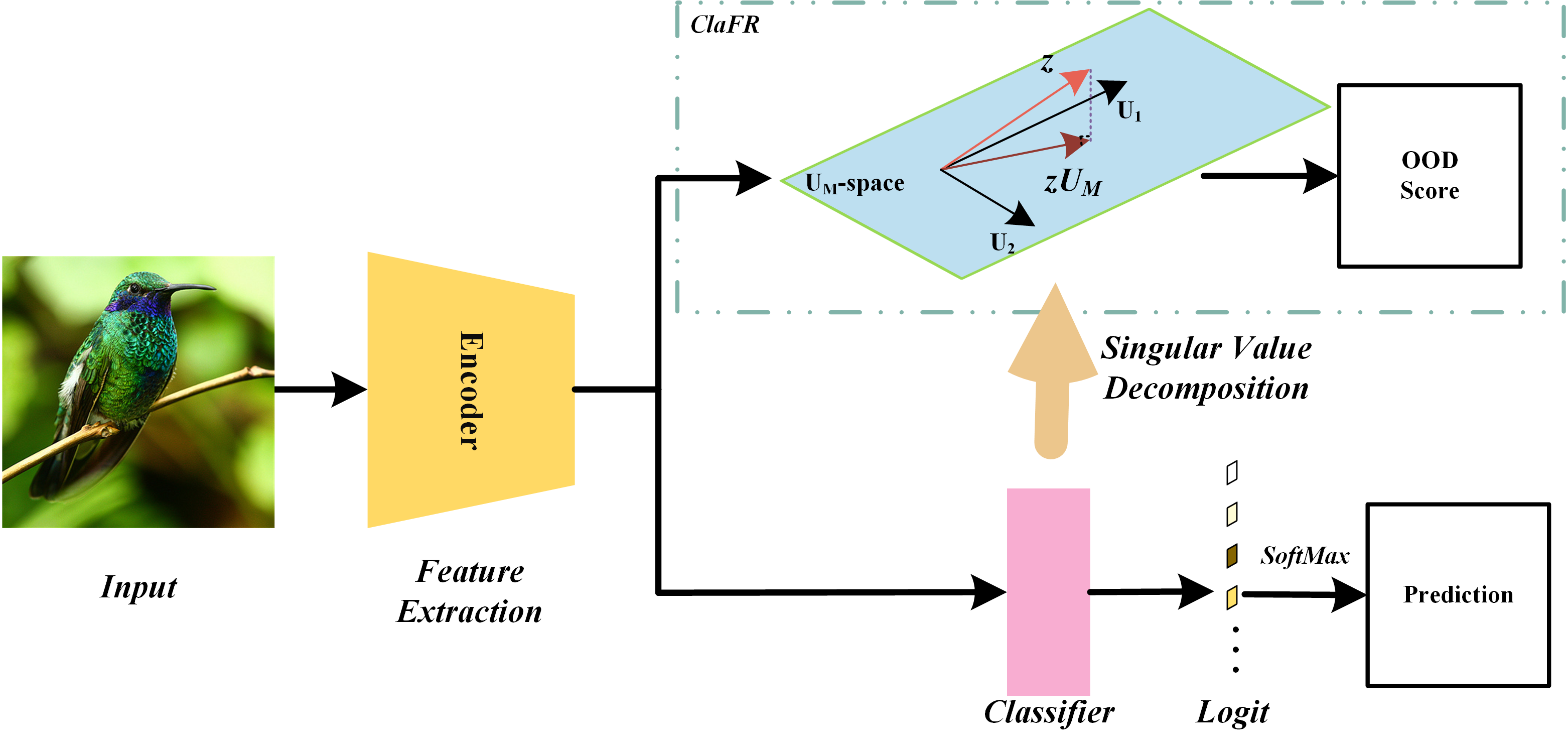}
  \caption{The overview of ClaFR.
  During inference, after extracting features through the pre-trained network, these features are projected into the subspace obtained by performing SVD decomposition on the classifier weights.
  Each sample's OOD score is calculated by ~\eqref{eq: score}.
  Generally, the lower the score, the more likely the sample is considered to be OOD.
  }
  \label{fig:example}
\end{figure}

Although training-agnostic methods do not require modifications to the original network architecture, they still rely on access to all training data for OOD detection, which is often impractical due to privacy concerns or sensitive data. Moreover, these methods typically demand significant computational resources. When training data is unavailable, an important question arises: how can we effectively leverage discriminative information in the feature space? Existing studies show that a model’s classifier holds class decision information. For example, in \cite{tanwisuth2021prototype}, the classifier's neural network weights are treated as class prototypes, while in \cite{qu2024lead}, the weight matrix is used to explore the feature space of the source data.

In this work, we follow the concept of feature decomposition, aiming to construct a feature subspace to enhance OOD detection performance.
This paper proposes a simple yet effective method called \textbf{Cla}ssifier-based \textbf{F}eature \textbf{R}econstruction (\textbf{ClaFR}).
ClaFR shifts the perspective towards the weights of the model classifier.
Specifically, ClaFR applies Singular Value Decomposition (SVD) to the final weight matrix to obtain a class-known subspace formed by singular vectors corresponding to larger singular values.
Subsequently, we project the latent representations from the penultimate layer into the discriminative subspace obtained from the weight matrix to create new data representations.
In this subspace, compared to OOD data, ID data exhibits smaller feature reconstruction errors, demonstrating greater robustness.
This is primarily because the feature distribution of OOD data is unknown to the model and significantly different from that of ID data, leading to a loss of key information during the feature reconstruction process for OOD data.

Compared to existing methods, ClaFR offers several attractive features.
Firstly, ClaFR does not rely on accessing training data to achieve competitive performance, making our algorithm well-suited for OOD detection scenarios that require data privacy protection.
Additionally, compared to methods that require large amounts of training data to learn OOD detectors, the subspace we learn avoids expensive computational costs and converges more quickly.
Our approach requires only lightweight access to the model's linear classification layer, allowing for fast updates to the OOD detector without the need to revisit the training database when the model is subsequently updated or new classes are added.

In summary, our \textbf{key contributions} are as follows:
\begin{itemize}
    
    \item We propose a novel OOD detection method termed CLAssifier-based Feature Reconstruction.
    ClaFR innovatively utilizes only the classifier's weight information to obtain class-known subspaces, without relying on training data features, thereby enhancing data privacy protection capabilities in the OOD detection domain.

    \item New insights into the class-known subspace reveal that it can capture and retain key information in ID data features, while OOD data cannot ensure this, thereby facilitating the detection of anomalous OOD data.

    \item ClaFR achieves optimal average performance on both large and small benchmarks, demonstrates strong generalization capability, and does not require heavy computational demands.

\end{itemize}

\section{Related Work}
\subsection{Out-of-Distribution Detection}
The key to OOD detection lies in assigning a score to each unknown sample to evaluate whether it belongs to the distribution observed during model training (ID), essentially forming a binary classification task~\cite{zhang2023openood, lu2025ood}.
As a baseline approach, MSP~\cite{hendrycks2016baseline} and ODIN~\cite{liang2017enhancing} suggest using the maximum softmax probability value as a measure of confidence. Energy~\cite{liu2020energy} quantifies the model's uncertainty about a sample by calculating the logsumexp on the model's output logits.
In contrast, Maxlogit~\cite{basart2022scaling}, considering that softmax might reduce the differences between ID and OOD samples, directly uses the highest logit value to measure the sample's uncertainty.
Additionally, KL-matching~\cite{basart2022scaling} calculates the divergence between softmax probabilities and class-conditional distributions.
Mahalanobis~\cite{lee2018simple, mueller2025mahalanobis++}, on the other hand, determines potential ID samples by calculating the Mahalanobis distance between test samples and class centroids, where a shorter distance indicates greater likelihood of being an ID sample.
ASH~\cite{djurisic2022ash} prunes anomalous neurons in the network using advanced threshold search techniques to recalibrate feature representations.
Methods based on logits and probability often overlook information in the weight matrix space, whereas feature-based methods require extensive computation.
Our approach effectively utilizes information from the weight matrix and achieves efficient OOD detection performance through lightweight computation.

\subsection{Feature Decomposition}
ID and OOD data often exhibit different patterns during feature decomposition or rectification~\cite{he2015cross, zhu2022boosting}, which can be leveraged for OOD detection.
For instance, NuSA~\cite{cook2020outlier} leverages this by decomposing features into two distinct components, then comparing the norm of the component specifically influencing OOD detection against the norm of the original input to effectively identify OOD instances. 
More recently, ViM~\cite{wang2022vim} combines information from both the feature space and logits, computing "virtual logits" from principal component residuals of the training data, which are subsequently integrated with an energy function to derive a comprehensive data score. 
Similarly, Neco~\cite{ammar2024neco} emphasizes the critical role of the principal component-relevant subspace in OOD detection, particularly within the context of neural collapse phenomena.
On the other hand, ~\cite{guan2023revisit} also employs PCA to process training data to obtain the subspace corresponding to the principal components, reconstructs data features through this subspace, and evaluates data confidence by computing the reconstruction error.

Although these methods are effective, they heavily depend on the availability of training data.
When training data is unknown, we define the class-known subspace through the classifier.
Our approach utilizes the mapping of this subspace to identify OOD data.

\section{Preliminaries}
For the sake of simplifying the expression of formulas, here we define some basic mathematical notation.
Let $\mathcal{X}$ = $\mathbb{R}^d$ and \( \mathcal{Y} = \{1, \ldots, C\} \) denote the input space and the corresponding label space, respectively. The joint distribution is denoted as \(D_{X Y} \) defined on $\mathcal{X} \times \mathcal{Y}$.
We denote the marginal probability distribution on $\mathcal{X}$ by $D_\mathcal{X}^{\text{ID}}$.
During testing, there exist some unknown OOD joint distributions \(D_{X Y^{'}} \) defined on \( \mathcal{X} \times \mathcal{Y}^{'} \), where
$ \mathcal{Y} \cap \mathcal{Y}' = \emptyset $.
Here, $x$ represents a sample data point.
Let $z$ denote the feature vector of the penultimate layer output corresponding to that sample.

During the deployment stage, the model needs to accurately identify when it encounters OOD data ($ y \in \mathcal{Y}^{'}$) that the data is from an unknown distribution and refuses to make predictions.
OOD detection can be considered as a binary classification task, determining whether a test image $x$ is an ID image.
Formally, OOD detection is expressed as follows:
\begin{equation}
G(x) = \left\{\begin{matrix}
 ID, & if~S(x) \geq \tau \\ 
 OOD, & if~S(x) <  \tau 
\end{matrix}\right.
\end{equation}
where $S(x)$ denotes a scoring function and $\tau$ is a threshold, which is usually chosen to ensure that a high proportion of ID data (e.g., 95\%) is correctly classified.

\begin{table*}[!t]
\setlength{\tabcolsep}{1.5mm}
\caption{For OOD detection comparison of ClaFR vs.
Baseline on the ImageNet-1k benchmark.
And the OOD datasets include SUN, iNaturalist, Places, Textures, and ImageNet-O.
We compare method performance using AUROC (AUC) and FPR95 (FPR) as evaluation metrics across ResNet-50 and MobileNet.
↑ indicates larger values are better, and vice versa.
All values are presented as percentages.
The best results are highlighted in bold, and the second are underlined.
}
\resizebox{\textwidth}{!}{%
\begin{tabular}{llcccccc}
\hline
\multirow{2}{*}{Model} & \multirow{2}{*}{Methods} & \textbf{SUN} & \textbf{iNaturalist} & \textbf{Places} & \textbf{Textures} & \textbf{ImageNet-O} & \textbf{Average} \\ 
 & & AUC↑ FPR↓ & AUC↑ FPR↓ & AUC↑ FPR↓ & AUC↑ FPR↓ & AUC↑ FPR↓ & AUC↑ FPR↓ \\ \hline
\multirow{11}{*}{ResNet-50}
 & MSP~\cite{hendrycks2016baseline} & 84.55 ~ 59.60 & 93.77 ~ 29.72 & 84.27 ~ 60.95 & 84.89 ~ 50.00 & 62.27 ~ 89.55 & 81.95 ~ 57.97 \\
 & Mahalanobis~\cite{lee2018simple} & 36.93 ~ 99.60 & 43.27 ~ 99.53 & 85.69 ~ \textbf{40.52} & 70.55 ~ 84.36 & 71.18 ~ 84.45 & 61.52 ~ 81.29 \\
 & Energy~\cite{liu2020energy} & 88.91 ~ 47.05 & 94.17 ~ 25.99 & 82.70 ~ \underline{51.14} & 88.89 ~ 39.32 & 70.38 ~ 85.35 & 86.01 ~ 50.77 \\
 & SSD~\cite{sehwag2021ssd} & 88.84 ~ 54.24 & 92.30 ~ 45.18 & 84.93 ~ 63.90 & 90.12 ~ \underline{20.26} & 77.45 ~ \textbf{65.60} & 86.73 ~ 49.83 \\
 & MaxLogit~\cite{basart2022scaling} & 88.42 ~ 50.97 & 92.98 ~ 30.46 & \underline{87.37} ~ 58.81 & 88.42 ~ 42.26 & 69.87 ~ 86.85 & 85.41 ~ 53.87 \\
 & KNN~\cite{sun2022out} & \underline{89.45} ~ \textbf{40.97} & 95.16 ~ 26.16 & 85.41 ~ 45.20 & 85.87 ~ 45.58 & 61.44 ~ 91.45 & 83.87 ~ 49.87 \\
 & KL-Matching~\cite{basart2022scaling} & 80.71 ~ 87.50 & 92.19 ~ 44.46 & 80.45 ~ 83.86 & 84.36 ~ 65.19 & 64.48 ~ 84.45 & 80.44 ~ 73.09 \\
 & ViM~\cite{wang2022vim} & 84.15 ~ 76.00 & 84.61 ~ 83.19 & 80.09 ~ 78.02 & \underline{92.71} ~ 35.72 & \underline{76.66} ~ \underline{69.25} & 83.64 ~ 68.43 \\
 & ASH~\cite{djurisic2022ash} & 89.15 ~ 47.20 & \underline{95.24} ~ \underline{22.69} & \textbf{87.72} ~ 51.42 & 89.30 ~ 37.85 & 70.88 ~ 84.65 & 86.46 ~ 48.76 \\
 & Neco~\cite{ammar2024neco} & 88.40 ~ 53.05 & 95.17 ~ \textbf{22.33} & 86.25 ~ 55.76 & 92.49 ~ 34.58 & 71.65 ~ 84.95 & 86.99 ~ 50.13 \\
 & ClaFR (ours) & \textbf{90.06}~ ~\underline{43.51} & \textbf{95.37} ~ 27.45 & 86.58 ~ 48.81 & \textbf{95.66} ~~\textbf{17.27} & \textbf{78.42} ~ 72.05 & \textbf{89.32} ~ \textbf{41.82} \\ \hline

\multirow{11}{*}{MobileNet}

 & MSP~\cite{hendrycks2016baseline} & 77.52 ~ 77.21 & 84.74 ~ 62.69 & 78.25 ~ 76.37 & 77.51 ~ 75.37 & 48.71 ~ 98.95 & 73.34 ~ 78.12\\
 & Mahalanobis~\cite{lee2018simple} & 22.77 ~ 99.52 & 29.26 ~ 99.37 & 26.01 ~ 99.23 & 74.05 ~ 67.83 & 77.51 ~ \textbf{70.70} & 45.92 ~ 87.33\\
 & Energy~\cite{liu2020energy} & 82.01 ~ 65.10 & 85.21 ~ 66.23 & 78.31 ~ 69.15 & 80.55 ~ 62.24 & 49.18 ~ 94.65 & 75.02 ~ 71.47\\
 & SSD~\cite{sehwag2021ssd} & 65.60 ~ 87.96 & 66.62 ~ 83.67 & 56.81 ~ 93.53 & 85.86 ~ 58.35 & \textbf{80.37} ~ 71.60 & 71.05 ~ 79.02\\
 & MaxLogit~\cite{basart2022scaling} & 85.71 ~ 61.27 & \underline{90.14} ~ 54.22 & 82.08 ~ 69.32 & 84.09 ~ 57.23 & 60.27 ~ 97.50 & 80.46 ~ 67.91\\
 & KNN~\cite{sun2022out} & 87.84 ~ \textbf{41.21} & 85.47 ~ 47.29 & 81.92 ~ 62.37 & 88.41 ~ 40.59 & 65.81 ~ 89.35 & 81.89 ~ 56.16\\
 & KL-Matching~\cite{basart2022scaling} & 76.35 ~ 81.16 & 88.63 ~ 54.21 & 76.32 ~ 79.14 & 87.55 ~ 80.26 & 57.19 ~ 95.60 & 77.21 ~ 78.07\\
 & ViM~\cite{wang2022vim} & 68.37 ~ 95.78 & 74.62 ~ 92.53 & 66.63 ~ 94.38 & \textbf{91.68} ~ \textbf{33.62} & \underline{78.24} ~ \underline{71.25} & 75.91 ~ 78.71\\
 & ASH~\cite{djurisic2022ash} & \underline{88.57} ~ 52.38 & \textbf{91.94} ~ \textbf{42.78} & \underline{82.28} ~ \underline{61.20} & 88.47 ~ \underline{34.04} & 62.99 ~ 93.65 & 82.85 ~ 56.81\\
 & Neco~\cite{ammar2024neco} & 84.18 ~ 67.84 & 89.29 ~ 59.71 & 80.64 ~ 73.66 & 85.47 ~ 56.19 & 61.43 ~ 97.45 & 80.20 ~ 70.97\\
 & ClaFR (ours) & \textbf{89.02}~ ~\underline{43.02} & 89.16 ~ \underline{46.28} & \textbf{82.62} ~ \textbf{55.72} & \underline{88.68} ~~42.01 & 68.71 ~ 84.06 & \textbf{83.63} ~ \textbf{54.21} \\ \hline
\end{tabular}%
}
\label{table:ImageNet for OODD}
\end{table*}

\section{Methodology}

In this section, we provide a detailed explanation of how ClaFR leverages SVD for subspace projection, including data processing and the scoring function.

\subsection{Classifier-based Feature Reconstruction}
Existing studies~\cite{tanwisuth2021prototype} have shown that the weights of the classification layer in deep neural networks are able to serve as proxies for the training data prototypes, thereby eliminating the need to access the training data.
We denote ${W} \in \mathbb{R}^{D \times C}$ as the weight matrix of the classifier, where $C$ represents the category number and $D$ represents the number of feature dimensions.
Given ${W}$, we perform SVD on it as follows:
\begin{align}
\begin{aligned}
    W &= U\varSigma V^T \\
    &= [U_M \, \vert \, U_{D-M}] \varSigma V^T \\
    &= [u_1, \dots, u_m \, \vert \, u_{m+1}, \dots, u_D] \varSigma V^T ,
\end{aligned}
\label{WSVD}
\end{align}
where \(\Sigma \in \mathbb{R}^{D \times C}\) is a diagonal matrix that contains the singular values, \(U \in \mathbb{R}^{D \times D}\) and \(V \in \mathbb{R}^{C \times C}\) are both orthogonal unitary matrices, $u_i$ is the i-th column vector of matrix $U$.

After performing SVD on matrix $W$, the singular values are arranged in descending order in the diagonal matrix $\varSigma$. Larger singular values indicate that there is more information in the corresponding directions, reflecting the primary structure and features of the data.
The left singular vector matrix $U$ provides an orthogonal basis for the output space.
We extract the first $m$ columns of matrix $U$ (i.e., $[u_1, \dots, u_m]$) to form the dimensionality reduction matrix $U_M \in \mathbb{R}^{D \times m}$ which is composed of the top-$m$ columns of the $U$ matrix after decomposition.

We retain sufficient information during dimensionality reduction to accurately approximate the original data by strategically selecting singular vectors that correspond to larger singular values. Regarding the determination of parameter m, our approach mirrors that of traditional dimensionality reduction techniques: we define m by choosing the top-m singular values that collectively account for approximately 90\% of the total sum of singular values, thereby ensuring a balance between dimensionality reduction and the preservation of essential data variance.

ClaFR aims to leverage the inherent structure within the classifier's weight subspace for robust OOD data identification, operating entirely without the need for external data resources. Building on previous research, our method focuses on the classifier's weight matrix \(W\), as it encapsulates the learned decision boundaries of known classes. We construct a class-known subspace, denoted as \(U_M\), by applying SVD to this weight matrix and selecting the singular vectors that correspond to the largest singular values. This approach is predicated on the understanding that, since the model is explicitly not trained on OOD data features, the constructed subspace inherently lacks critical information pertaining to these unseen OOD characteristics, thereby enabling effective discrimination.

We regard the subspace \(U_M\) as an orthogonal basis for the output space, providing a new coordinate system for feature mapping. By projecting the original data features onto the subspace \(U_M\), we obtain new data representations. For ID data, this projection operation retains more important components of the subspace, while for OOD data, it results in more information loss. Due to the orthogonality of the column vectors in the \(U_M\) matrix, as ensured by SVD decomposition, the mapping through \(U_M\) allows us to separate independent directions and reduce correlation among features, thus addressing the problem of high inter-feature correlation.

\begin{algorithm}[t]
    \label{pseudocode}
    \caption{Pseudo code of ClaFR.}
    \KwIn{Penultimate layer features $ z $.}
    \KwIn{Linear classifier weights $W$.}
    \KwIn{Cumulative Explained Variance $\alpha$.}

    $\hat{z} \gets \frac{z}{\|z\|_2}$ \tcp*{Normalize features}
    $W = U\Sigma V^T$ \tcp*{SVD of $W$}
    $\sigma_i = \Sigma[i]$ \tcp*{Singular value $\sigma_i$ from $\Sigma$ matrix}
    $m \gets {\textstyle \sum_{i=1}^{m}\sigma_i} > \alpha {\sum_{i=1}^{D} \sigma_i}$\ \tcp*{Number of m}
    Extract the top $m$ columns of $U$ to obtain $U_M$\
    \tcp*{To obtain the class-known subspace}
    $S(x) = || zU_M ||_2$\  \tcp*{OOD Score}
    \Return $S(x)$
\end{algorithm}

\subsection{OOD Score}
Now we need to define an appropriate OOD scoring function. In the inference phase, given a test sample \(x\) and its penultimate layer feature \(z\) from a well-trained DNN, 
The feature reconstruction error is expressed by the following formula:
\begin{align}
    \begin{aligned}
        e(x) &= -|| zU_M U_M^T - z ||_2 \\
        &= -\sqrt[]{||z||_2^2-||zU_M||_2^2}.
    \end{aligned}
    \label{eq:recon}
\end{align}
The calculation here considers the determined \( U_M \) as an orthogonal coordinate system.
By projecting the sample features onto the subspace \( U_M \) to obtain a low-dimensional representation of the data.
Then, by using the transpose of \( U_M \), the features are mapped back to the original feature space, completing the feature reshaping process.
Eq. \eqref{eq:recon} ultimately calculate the L2 norm of the difference between the features before and after reconstruction to measure the uncertainty of the sample.
Upon expanding the formula, we find that this reconstruction error calculates the negative projection distance of the vector \( z \) to the subspace \( U_M \).

Since the feature \( z \) and the subspace \( U_M \) are known, to simplify the calculation, we directly compute the projection length of feature \( z \) to the subspace:
\begin{equation}
    S(x) = ||zU_M||_2.
\label{eq: score}
\end{equation}
Based on the previous analysis, \(U_M\) is regarded as the principal component of ID data features.
When ID data is reconstructed in a low-dimensional manifold, it tends to be closer to the hyperplane of the known feature space $U_M$, which means the feature projection vector $zU_M$ has a larger vector norm~\cite{fang2024kernel}.
In contrast, OOD data, which the model has not encountered during training, shows smaller components in the projection, resulting in a score $S(x)$ that is usually lower than that of ID data.
Eq. \eqref{eq: score} is another representation of the feature reconstruction error, and it is more concise and efficient.
Finally, we use \( S(x) \) as our final scoring function.

\section{Experiment \& Results}

\subsection{Experiment settings}
In this section, we compare our method with current advanced approaches. 
Following previous work, we conduct comparisons across multiple OOD detection methods on both large-scale and small-scale datasets.
We consider ImageNet-1K~\cite{deng2009imagenet} (large-scale) and CIFAR-10 and CIFAR-100~\cite{krizhevsky2009learning} (small-scale) as the ID datasets.
In addition, an extensive ablation study provides further insights into the selection of subspace dimensions.

\textbf{Baselines.}
In our experiment, we conducte a comprehensive evaluation by comparing ClaFR with representative approaches.
all of which do not require modifications or fine-tuning of network parameters.
We selecte ten baseline methods, including five output-based methods: MSP~\cite{hendrycks2016baseline}, Energy~\cite{liu2020energy}, Maxlogit~\cite{basart2022scaling}, KL-Matching~\cite{basart2022scaling}, and six feature-based methods: Mahalanobis~\cite{lee2018simple}, SSD~\cite{sehwag2021ssd}, KNN~\cite{sun2022out}, ViM~\cite{wang2022vim}, ASH~\cite{djurisic2022ash}, Neco~\cite{ammar2024neco}.
Our experimental setup followes the configuration recommended by previous work, and the related baseline methods were consistent with those in the original paper.

\textbf{Evaluation Metrics.}
In alignment with the experimental details outlined in prior work~\cite{sun2021react, peng2024conjnorm}, we employ two widely recognized metrics for OOD detection to present our results.
First, we consider the FPR@95, which represents the false positive rate when the true positive rate (TPR) reaches 95\%.
A lower FPR@95 value signifies better OOD detection performance, as it indicates fewer incorrect predictions at a high true positive rate.
Second, we utilize the threshold-independent AUROC metric, which calculates the area under the receiver operating characteristic curve.

An AUROC value of 0.5 reflects a model whose classification performance is no better than random guessing, implying that its ability to separate OOD from in-distribution samples is negligible, while an AUROC value of 1.0 denotes a model with perfect classification capability, indicating complete and accurate discrimination.
For consistency and ease of interpretation, we report all results in percentage form.

\begin{table*}[t]
\setlength{\tabcolsep}{1.5mm}
\caption{For OOD detection comparison of ClaFR vs.
Baseline on the CIFAR benchmark.
The table has CIFAR-10 and CIFAR-100 as the ID datasets on the left and right sides (not displayed), and it shows the corresponding OOD datasets.
All values are presented as percentages.
The best results are highlighted in bold.
}
\resizebox{\textwidth}{!}{%
\begin{tabular}{lccc|ccc}
\hline
\multirow{2}{*}{Methods} & \textbf{CIFAR-100} & \textbf{SVHN} & \textbf{Average} & \textbf{CIFAR-10} & \textbf{SVHN} &\textbf{Average} \\ 
 & AUC↑ FPR↓ & AUC↑ FPR↓ & AUC↑ FPR↓ & AUC↑ FPR↓ & AUC↑ FPR↓ & AUC↑ FPR↓ \\ \hline
MSP~\cite{hendrycks2016baseline} & 91.51 ~ 44.98 & 95.62 ~ 15.24 & 91.57 ~ 30.11 & 74.83 ~ 85.76 & 84.76 ~ 65.73 & 78.62 ~ 75.46 \\
Mahalanobis~\cite{lee2018simple} & 56.91 ~ 93.59 & 95.77 ~ 48.51 & 74.84 ~ 71.05 & 59.49 ~ 98.42 & 81.32 ~ 78.20 & 70.40 ~ 88.31  \\
Energy~\cite{liu2020energy} & 92.37 ~ 39.59 & 97.88 ~ 12.43 & 95.12 ~ 26.01 & 72.16 ~ 84.75 & 83.24 ~ 58.75 & 77.70 ~ 71.75  \\
SSD~\cite{sehwag2021ssd} & 91.44 ~ 38.59 & 96.89 ~ 10.09 & 94.32 ~ 26.34 & 74.91 ~ 92.62 & 86.01 ~ 47.18 & 80.46 ~ 69.90 \\
MaxLogit~\cite{basart2022scaling} & 92.16 ~ 39.81 & 97.66 ~~~ 8.32 & 94.96 ~ 24.06 & 74.18 ~ 82.05 & 84.64 ~ 62.09 & 79.41 ~ 72.09 \\
KNN~\cite{sun2022out}  & 92.94 ~ 41.07 & 98.33 ~ 12.35 & 95.35 ~ 26.71 & 71.78 ~ 88.31 & 89.50 ~ 56.08 & 80.64 ~ 72.19 \\
KL-Matching~\cite{basart2022scaling}  & 88.15 ~ 47.52 & 92.13 ~ 20.36 & 90.14 ~ 33.94 & 74.24 ~ 86.93 & 82.96 ~ 85.22 & 78.60 ~ 86.07 \\
ViM~\cite{wang2022vim} & 90.58 ~ 56.53 & 99.02 ~~~ 5.09 & 95.30 ~ 30.81 & 73.08 ~ 92.57 & 88.94 ~ 46.78 & 81.01 ~ 69.68 \\
ASH~\cite{djurisic2022ash} & 93.30 ~ 39.91 & 98.67 ~~~ 6.48 & 95.98 ~ \textbf{21.19} & 74.99 ~ 85.64 & 89.15 ~ 58.97 & 82.57 ~ 71.81 \\
Neco~\cite{ammar2024neco} & 93.26 ~ 39.74 & 98.77 ~~~ 5.67 & 96.01 ~ 22.70  & 76.02 ~ 84.01 & 88.59 ~ 61.26 & 82.20 ~ 72.63 \\
ClaFR (ours) & 93.62 ~ 39.95 & 98.96 ~~ 7.04 & \textbf{96.29} ~ 23.49 & 75.29 ~ 87.51 &  90.36 ~ 49.31 & \textbf{82.82 ~ 68.41} \\ \hline
\end{tabular}%
}
\label{table:CIFAR for OODD}
\end{table*}
\textbf{Datasets.}
Following the research setup of prior work, we primarily focus on OOD detection performance in large-scale image classification tasks.
Our experiments on the ImageNet benchmark demonstrate the effectiveness of our method.
For experiments involving ImageNet-1K as the ID dataset, we evaluate the model's performance using our method on five widely recognized OOD benchmark datasets: SUN~\cite{xiao2010sun}, iNaturalist~\cite{van2018inaturalist}, Places~\cite{zhou2017places}, Textures~\cite{cimpoi2014describing}, and ImageNet-O~\cite{wang2022vim}, which is a sufficiently challenging OOD dataset.
The five tested OOD datasets do not have category overlap with the ImageNet-1K, and their categories involve different semantic information.
The performance on these five datasets can typically indicate the robustness and scalability of OOD detection methods.
During the inference phase, all test data are resized to 224×224 following the standard processing procedure.
In addition, we also evaluate the effectiveness of ClaFR on small-scale datasets. Specifically, for experiments using CIFAR-10 (resp. CIFAR-100) as the ID datasets, we utilize CIFAR-100 (resp. CIFAR-10) and SVHN~\cite{netzer2011reading} as the OOD datasets.
Due to space limitations, we report only their average results.

\textbf{Models Settings.}\label{AA}
We conduct experiments using ResNet-50~\cite{he2016deep} and MobileNet~\cite{sandler2018mobilenetv2} as backbones on ImageNet-1k.
On the ImageNet-1k dataset, it is trained from scratch using contrastive loss to enhance performance.
This approach encourages samples of the same class to be closer in the embedding space while maximizing the distance between samples of different classes, achieving a classification accuracy of 78.12\% on this dataset.
Another network is MobileNet, which is an efficient, lightweight convolutional neural network model designed for mobile and embedded vision applications.
In our experiments, we use the officially released pre-trained weights of the MobileNet-v2 version, which introduces inverted residual structures and a linear bottleneck.
For the CIFAR benchmark, ResNet34 serves as the backbone for feature extraction.
The model is fine-tuned under the supervised contrastive learning framework.
, across 1,000 epochs, separately on the CIFAR-10 or CIFAR-100 datasets.
When evaluated on CIFAR-10, the network achieves a classification accuracy of 95.28\%, while it records an accuracy of 73.23\% on CIFAR-100.
Standard dataset splits are used for these evaluations, comprising 50,000 training images and 10,000 test images.

\textbf{Implementation details.}
For comparison, we keep the hyperparameters of various methods consistent with those in the original literature.
Similar to the dimensional selection in ViM and Neco, we adhere to the configurations specified in the source code.
It is also noteworthy that for the Neco method, its primary results are obtained under specific conditions, namely the setting of neural collapse.
This implies that the original model must undergo 1000 epochs of fine-tuning on the ImageNet-1K dataset in a state of overfitting to derive the experimental results.
It is worth noting that, among the various methods mentioned in ASH~\cite{djurisic2022ash}, we selecte only the optimal ASH-P version for our comparative experiments.
The dimensions of the weight matrix \(D\) are consistent with the data features.
ResNet-50 corresponds to a dimensionality of 2048, MobileNet corresponds to 1280, and ResNet34 corresponds to 512.
In selecting the subspace dimensionality, we determine this by accumulating the proportion of singular values with respect to the total singular value using a hyperparameter \(\alpha\), which is typically set to 90\%.
For specific ablation experiments on dimension selection, please refer to Fig.~\ref{fig:ablation}.

\subsection{Results Analysis}

\textbf{Results on ImageNet-1k.}
In Table \ref{table:ImageNet for OODD}, we compare the performance of advanced OOD detection methods on the ImageNet-1k dataset.
These methods do not require modifications to existing neural network structures or parameters and can detect OOD data without compromising the model's classification capability.
From the results, ClaFR achieves state-of-the-art average performance across five commonly used representative OOD detection datasets.
With the ResNet-50 backbone, ClaFR achieves an average of 89.32\% AUROC and 41.82\% FPR95, representing improvements of 2.33\% in AUROC and 6.94\% in FPR95.
In the lower part of the table, using the MobileNet network as an example, our results remain leading, with improvements of 0.88\% in AUROC and 1.95\% in FPR95 compared to the second-best results.
Under these comprehensive and challenging OOD data conditions, our method demonstrates competitive OOD detection performance with lightweight computational requirements.

\textbf{Results on CIFAR.}
To demonstrate the effectiveness of ClaFR in detecting OOD data on small-scale datasets and models, Table \ref{table:CIFAR for OODD} presents the experimental results using CIFAR10 / CIFAR-100 as the ID dataset on the ResNet34.
Due to space constraints, we only report the average results on OOD data.
Refer to the supplementary materials for detailed experimental results.
The left side of the table shows the average performance when using CIFAR-10 as the ID dataset, with CIFAR-100 and SVHN as the OOD datasets.
Conversely, the right side uses CIFAR-100 as the ID dataset, with CIFAR-10 and SVHN as the OOD datasets.
Overall, on the CIFAR Benchmark, ClaFR exhibits competitive performance compared to existing SOTA methods and maintains a leading position in terms of overall performance.

\subsection{Ablation Experiment}
\label{blation}
\begin{figure}[h!]
  \centering
  \includegraphics[trim=5 5 5 5, clip, width=\linewidth]{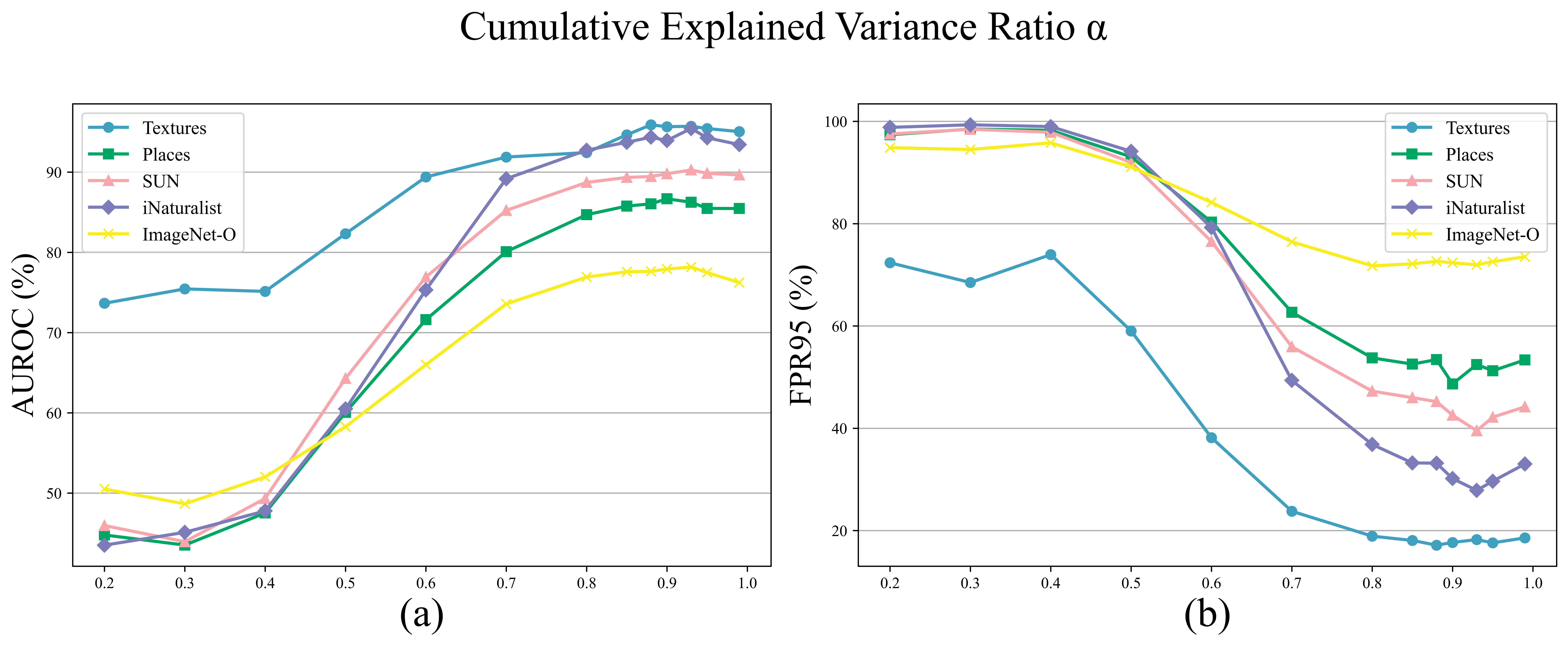}
  \caption{
    An ablation study on the sensitivity of the cumulative explained variance ratio \( \alpha \) is conducted to veri its impact on OOD detection performance.
  }
  \label{fig:ablation}
\end{figure}

Our approach involves applying SVD to the model's weight layers to obtain a discriminative subspace.
In this process, the only hyperparameter we need to determine is the choice of subspace dimension.
According to machine learning theory, we use the cumulative explained variance ratio, \( \alpha \), for ablation experiments.
Specifically, \( \alpha \) represents the proportion of energy contained in the top-\( m \) singular values relative to the total energy of all singular values after SVD decomposition.
The hyperparameter $\alpha$ determines the extent to which data features are retained and is used to balance the differences between reshaped features of ID data and OOD data.
On the ResNet50 backbone using the ImageNet-1k benchmark, five standard OOD datasets are evaluated.
The results of the ablation experiments are shown in the Fig. \ref{fig:ablation}.
The bottom horizontal axis in the figure represents the proportion of original singular values retained.

Our experimental results indicate that the hyperparameter \(\alpha\) yields competitive results over a wide range of values, suggesting that ClaFR is not overly sensitive to the choice of \(\alpha\).
When the selected proportion of singular values \( \alpha \) is approximately 80\% to 95\%, the model consistently achieves a relatively high rate of accurately identifying OOD data. This corroborates our hypothesis that in such a discriminative subspace, ID data features demonstrate increased robustness.
Conversely, OOD data experiences a substantially higher rate of information loss when mapped through such a subspace compared to ID data, resulting in a significant degradation of data features that hinders effective reconstruction.

\subsection{The low complexity of ClaFR}
In current scenarios, especially when handling large-scale datasets, existing methods usually require a large amount of labeled training data features to train OOD detectors.
This is challenging in practical applications, as any changes in the network architecture post-deployment or expansion of the training dataset necessitate retraining the OOD detector, resulting in significant computational time and memory consumption.
All experiments are conducted on a single NVIDIA GeForce RTX 3090 GPU.

\begin{table}[h]
\centering
\caption{Comparison of the KNN algorithm in terms of complexity, inference time, and memory consumption.}
\begin{tabular}{c|c|c|c}
\hline
\textbf{method} & \textbf{complexity} & \textbf{Inference (ms, per sample)} & \textbf{storage} \\ \hline
KNN & $\mathcal{O}(N_{tr})$ & $\approx 8.26$ & $\approx 20 \, \text{GiB}$ \\
ClaFR & $\mathcal{O}(1)$ & $\approx 0.012$ & $\approx 12 \, \text{MiB}$ \\ \hline
\end{tabular}
\label{complexity}
\end{table}
In contrast, ClaFR does not require fine-tuning of training data features, but instead makes lightweight adjustments to the weights of the model's classification layer.
In Table \ref{complexity}, we compare the time complexity and inference time of our method with existing methods, and the results show that our method has significantly faster inference speed.

\section{Conclusion}
In this paper, we propose Classifier-based Feature Reconstruction (ClaFR), a novel OOD detection algorithm.
Our algorithm models the OOD detection task by incorporating information from the weights of the linear classification layer.
ClaFR utilizes SVD to extract class-known subspaces from these weights, and then projects data features into these subspaces to obtain new data representations.
It distinguishes anomalous data by measuring the reconstruction error of these new representations.
We conducted extensive experiments on both large-scale and small-scale OOD benchmarks, utilizing a diverse range of models.
Our evaluation results highlight the superiority and robustness of ClaFR for OOD detection. We also provide theoretical insights, demonstrating its effectiveness through subspace projection compared to other methods.
Notably, ClaFR does not require access to the training database, making it ideal for privacy-sensitive scenarios.
Furthermore, it is computationally efficient, requiring only minimal computational and memory resources, thereby making it a practical solution for OOD detection tasks.

\section*{Acknowledgements}
This work was funded by the National Natural Science Foundation of China under Grants (62276256, U2441251), the Young Elite Scientists Sponsorship Program by CAST (2023QNRC001), and the Young Scientists Fund of the State Key Laboratory of Multimodal Artificial Intelligence Systems (ES2P100117).

\bibliographystyle{splncs04}
\bibliography{main}

\end{document}